%% file: main.tex
\documentclass[conference]{IEEEtran}
\newtheorem{Definition}{Definition}

\IEEEoverridecommandlockouts
\usepackage{hyperref}
\usepackage{cite}
\usepackage{amsmath,amssymb,amsfonts}
\usepackage{algorithmic}
\usepackage{algorithm}
\usepackage{textcomp}
\usepackage{mwe}
\usepackage{multirow, multicol}
\usepackage{booktabs}
\usepackage{tabularx}
\usepackage{diagbox}
\usepackage{float}   
\usepackage[framemethod=tikz]{mdframed}
\usepackage{psfrag,epsfig,url,fancyhdr,color}
\usepackage{setspace}
\usepackage{mathtools}
\usepackage{graphicx}
\usepackage{xcolor}
\usepackage{caption}
\usepackage{subcaption}

\newcommand{\etal}{{\it et al.}}

\def\BibTeX{{\rm B\kern-.05em{\sc i\kern-.025em b}\kern-.08em
    T\kern-.1667em\lower.7ex\hbox{E}\kern-.125emX}}

\DeclareMathAlphabet{\mathdata}{OMS}{cmsy}{m}{n}

\begin{document}

\pdfoutput=1
\title{DG-Trans: Dual-level Graph Transformer for Spatiotemporal Incident Impact Prediction on Traffic Networks}

\author{\IEEEauthorblockN{Yanshen Sun\textsuperscript{1}, Kaiqun Fu\textsuperscript{2}, and Chang-Tien Lu\textsuperscript{1}}
\IEEEauthorblockA{%\textit{dept. Computer Science} \\
\textit{\textsuperscript{1}Virginia Tech}, \textit{\textsuperscript{2}South Dakota State University}\\
% Falls Church, VA\\
yansh93@vt.edu, kaiqun.fu@sdstate.edu, ctlu@vt.edu}\\
}
\IEEEaftertitletext{\vspace{-2\baselineskip}}
\maketitle

\begin{abstract}
The prompt estimation of traffic incident impacts can guide commuters in their trip planning and improve the resilience of transportation agencies' decision-making on resilience. However, it is more challenging than node-level and graph-level forecasting tasks, as it requires extracting the anomaly subgraph or sub-time-series from dynamic graphs. In this paper, we propose DG-Trans, a novel traffic incident impact prediction framework, to foresee the impact of traffic incidents through dynamic graph learning. The proposed framework contains a dual-level spatial transformer and an importance-score-based temporal transformer, and the performance of this framework is justified by two newly constructed benchmark datasets.
The dual-level spatial transformer removes unnecessary edges between nodes to isolate the affected subgraph from the other nodes. Meanwhile, the importance-score-based temporal transformer identifies abnormal changes in node features, causing the predictions to rely more on measurement changes after the incident occurs. Therefore, DG-Trans is equipped with dual abilities that extract spatiotemporal dependency and identify anomaly nodes affected by incidents while removing noise introduced by benign nodes. Extensive experiments on real-world datasets verify that DG-Trans outperforms the existing state-of-the-art methods, especially in extracting spatiotemporal dependency patterns and predicting traffic accident impacts. It offers promising potential for traffic incident management systems.

\end{abstract}


\input{sections/introduction}

\input{sections/relatedwork.tex}
\input{sections/problem_def}
\input{sections/method_copy.tex}
\input{sections/experiment_copy.tex}
\input{sections/conclusion}

\bibliographystyle{IEEEtran}
\bibliography{main}

\end{document}

%% file: sections/introduction.tex
\section{Introduction}

The efficient detection of non-recurring congestion caused by traffic incidents is
of great importance for modern intelligent transportation systems
(ITS). Furthermore, the estimation of the impact of such incidents is crucial
due to the potential social and economic loss~\cite
{adler2013road}. However, effectively pinpointing traffic incidents is challenging due to high data volatility, the lack of incident labels,
and the demands of ultra-low inference times in modern ITS. Despite this
difficulty, traffic incident detection and diagnosis is an active research
discipline, including approaches related to distributed computing, transportation
management, and urban resource management. The widespread deployment of traffic
sensors and traffic incident management systems (TIMS) has promoted such research by generating enormous
amounts of high-dimensional traffic sensor records and making them ubiquitously
accessible. With the abundance of traffic data sources, we can now present
two datasets and an efficient neural network model for accurate traffic
incident impact prediction from both the spatial and temporal perspectives.
\begin{figure*}[t]
  \centering
  \includegraphics[width=\linewidth]{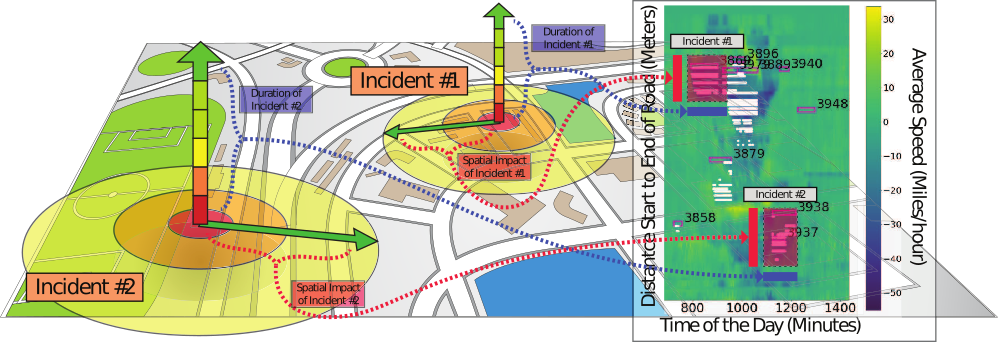}
  \caption{\textbf{An example of traffic incident impact.} The impact of incidents can be identified by congestion. The closer a location is to the incident, the longer it is impacted. Two factors, impact length (spatial impact) and duration (temporal impact), are used to define an incident's impact. In the map (left), the vertical arrows and the circles indicate congestion alleviation over time and distance, respectively. In the speed heatmap (right), the white dots reflect sensed congestion related to incidents.}
  \label{fig:intro_case}
\end{figure*}
The impact of traffic incidents can be quantified in two dimensions: time and space. As illustrated in Figure~\ref{fig:intro_case}, traffic along a road can be plotted over time and location as a heatmap of speed. The impact of an incident usually presents itself as a drop in speeds, that starts at the time and location of the incident and then spreads upstream before finally peaking. Conventionally, the incident duration is defined as the time between the validation and restoration of an incident, while the impact length is the number of cars blocked by the incident. However, we observe that cars are forced to slow down even if they are not near the incident location. In this case, for the spatial dimension, we redefine the impact length on the road as the maximum continuous congestion distance immediately upstream of the incident.

Recent research has primarily focused on the temporal impact of incidents~\cite{fu2019titan, fu2021hierarchical, meng2022early}, with limited attention paid to their spatial impact. Traffic forecasting studies have developed models for learning spatiotemporal representations. These could be leveraged to assess incident effects, though their abilities in sub-graph extraction and summarization remain untested. Based on the previous work above, we can conclude that the current research has the following drawbacks.

\textbf{The spatiotemporal quantification of traffic incident impacts is rarely considered, and the publicly available supported data is limited.} Unlike traditional transportation research, detailed traffic data -- such as the number of cars in the congestion queue and traffic signal states -- are usually unavailable to the public. In this case, it is essential to define extensive ``traffic incident impact'' criteria and construct open-source datasets in the context of dynamic graph data mining. 

\textbf{Relations between the static road network and the dynamic traffic correlation network are not properly modeled.} 
% Generally, spatial information aggregation can be considered as updating a node's features with the features of its neighbors, while the effect of each neighbor node is
%  determined by the weights on the edges. 
% Previous studies often incorporate human knowledge into latent feature correlations by utilizing one or more pre-computed adjacency matrices to mask attention weights~\cite{xu2020spatial,wu2021representing,xu2020spatial}. However, they rarely consider the errors that can be introduced by attention parameters or pre-computed weight matrices. 
In traffic forecasting scenarios, in particular, distance-based adjacent matrices are not always accurate, and sensor measurements may not provide sufficient information for similarity estimation. Therefore, it is necessary to develop a new approach that takes full advantage of the flexibility of attention and emphasizes the importance of road network structure.

\textbf{The capabilities of dynamic graph learning models are not evaluated for task-oriented subgraph/sub-time-series extraction.} In a traffic loop sensor network, an incident can dramatically affect a subset of sensors for a limited period. Therefore, it is necessary to develop dynamic graph learning models that can accurately predict the impacts of such incidents. 
% These models should be able to extract relevant sensor readings and the mutual effects of the incident on the affected sensors. 
Despite the prevalence of node and graph-level regression/classification tasks in dynamic graph representation learning, the ability to predict incident impacts using such networks has not been fully explored.

% The effect of incidents varies case by case. A ``road maintenance'' case can affect traffic patterns for days and long ranges, while a ``rear-end collision'' at midnight usually causes no or only a small range of traffic congestion. Congestion may form before the validation time during peak hours, and may also take some time to gather enough cars during non-peak hours. Besides, different from general graph-level prediction tasks, incident impact prediction is not essentially a summarization of the entire network. An incident usually only causes variation in a subgraph, while the spatiotemporal coordinates of the variation are hard to determine. Thus, the spatiotemporal relations between sensors and incidents are complex. 

To solve these challenges, we propose the \textbf{D}ual-level \textbf{G}raph \textbf{Tran}sformer for Traffic Incident Spatiotemporal Impact Prediction. Our approach effectively extracts spatiotemporal relations within the
traffic network and automatically locates latent variations in traffic
patterns with two benchmarks measuring the impacts of incidents. The main
contributions of this paper are summarized as follows:

\begin{itemize}
    \item \textbf{Quantifying the definition of ``incident impact'' and providing two open-source datasets.} We provide new definitions of traffic incident impact based on the observations of the spatiotemporal relationships between incident records and sensor measurements. As a result, the incident impact problem can be studied using dynamic graph representation learning methods. In addition to the sensor network and incident records, we also provide auxiliary data, such as sensor metadata, road structures, and other data, in order to support broader applications and problem domains.
    
    \item \textbf{Proposing an efficient dual-level graph attention strategy utilizing accurate connectivity and correlation information.} The proposed graph attention module integrates the advantages of static graph structures and graph attention. As a result, it properly guides attention weights with static graph structure knowledge, making full use of the flexibility of attention while stressing the importance of road network structure. In addition to its superior performance in extracting traffic patterns, our attention module is also proved to be time-efficient.
     
    \item \textbf{Designing an importance score temporal transformer mechanism to locate sub-sensor-graph affected by incidents.}
     Inspired by TranAD~\cite{tuli2022tranad}, we computed an ``importance score'' for each sensor at each timestamp using anomaly detection techniques. As a higher variation in traffic patterns leads to a larger ``importance score'', sensors potentially affected by an incident can be accurately located and assigned higher weights during prediction.
     
    \item \textbf{Evaluating model performance with extensive
     experiments.} Overall, DG-Trans outperforms state-of-the-art traffic prediction models on the two datasets we provided. We also show that each module in DG-Trans is  effective. The efficiency of our spatial transformer module is proved mathematically. The ``importance score'' is also proved to be effective through our case study. We also explored different methods of incorporating the metadata into the traffic network to lay the groundwork for future research.
\end{itemize}

This article is structured as follows. We first introduce existing traffic incident impact prediction studies and dynamic graph representation learning models in Section~\ref{sec:related}, then provide the definition of traffic incident impact prediction problem. In Section~\ref{sec:method}, we elaborate on our proposed model. Section~\ref{sec:experiment} describes our benchmark datasets and shows the advantages of our model.
% , as well as other models's limitations. 
Finally, Section~\ref{sec:conclusion} summarizes this work.

%% file: sections/relatedwork.tex
\section{Related Work} \label{sec:related}

Traffic incident impact prediction has been considered an important topic in traffic management for decades. Traditional studies considered it as a 1-D propagation task and designed deterministic queuing diagrams~\cite
{Ben2001network} and shockwave theory~\cite{motamedidehkordi2016shockwave}.~\cite{tang2020statistical} summarizes statistical and machine learning methods arguing that quantile regression (QR), finite mixture (FM), and random parameters hazard-based duration (RPHD) perform the best.
% A queuing model is constructed to measure the queue lengths and wait times. The traffic shockwave theory determines the delay time by measuring the difference between the average traffic flow and density. 

\textbf{Grid-based Representation Learning.} With the development of machine learning methods, more complex factors are considered in traffic incident impact prediction models. For example, Zou~\etal~\cite{zou2021application} utilize Bayesian Model Averaging (BMA) to merge the prediction results of multiple machine learning models. Meanwhile, Match-Net~\cite{kalair2021dynamic} gradually updates its prediction as new data is acquired and performs interpretability analysis using Shapley values. 

%% more about analyzing accident records, no need to be included
%~\cite{GRIGOREV2022incident} provides three traffic incident datasets. The datasets contain auxiliary information and support two incident-duration-based tasks: long/short-term classification and incident duration prediction.~\cite{zhang2022analysis} labels traffic incidents with speed-changing rates and analyzes how road attributes affect congestion durations.~\cite{li2021online} defines incident post-impact prediction as a task performing classification, duration prediction, and accumulative queue length prediction.~\cite{lee2022quantifying} discusses the relations between incident report time and congestion occurring time. 

Some studies consider traffic data with regular data structure.
Huang~\etal~\cite{huang2020using} utilize a generative adversarial network (GAN) to predict incident impacts by directly learning speed heatmaps. Lin~\etal~\cite{lin2020real} gradually update incident duration predictions with a decision tree as new speed measurements are acquired. Similarly, Zhu and Wu~\cite{zhu2021dynamic} propose a model that updates incident duration prediction every five minutes. However, these models fail to consider the complexity of road networks, instead considering only one road.

\textbf{GNN-RNN Dynamic Graph Representation Learning.} Some researchers consider traffic forecasting as a downstream task of dynamic graph representation learning. For example, RadNet~\cite{tuli2022radnet} considers incident prediction as an anomaly detection problem and utilizes a transformer and GCN.
 % However, this study did not include experiments on real incident record data.
 PrePCT~\cite{bai2021prepct} proposes a new method of mapping road networks to grid cells and utilizes a CNN and LSTM for congestion prediction. DIGC-Net~\cite{xie2020deep} predicts traffic flow speeds by considering incident records and similarities among flow segments in the same time window. SLCNN~\cite{zhang2020spatio} utilizes static and dynamic graphs and top-K attention to perform a C3D-like~\cite{tran2015learning} temporal convolution. Finally, Yoo~\etal~\cite{yoo2021conditional} propose a covariance loss considering both the basis function space and the targeted variable space. 

\textbf{Attentive Dynamic Graph Representation Learning.} The attention mechanism is now one of the most popular methods of learning relations between graph nodes through their feature similarities. STAWnet~\cite{tian2021spatial} utilizes attention to extract spatial relations among road segments and employs gated TCN to extract temporal dependencies. 
% ADN~\cite{drakulic2022structured} merges the spatial and temporal dimensions with attention across all elements. Tygesen~\etal~\cite{tygesen2022unboxing} perform traffic prediction with neural relational inference techniques. 
Other studies have utilized transformers for spatiotemporal
 data mining. ~\cite{wu2021representing, rong2020self, zhang2020graph} consider a GNN as an auxiliary module for transformers. Except for GNN, graph structure and position features can also be utilized by transformers through position encoding~\cite{xu2020spatial, zhang2020graph,
 ying2021transformers}. For example, GraphGPS~\cite{rampavsek2022recipe} encodes graph structures and node positions in three different ways. ~\cite
 {dwivedi2020generalization, min2022masked} considers adjacent matrices as masks of attention. GraphiT~\cite{mialon2021graphit} turns an adjacent matrix into a kernel matrix. Other dynamic graph representation learning
 models are also efficient in traffic flow forecasting tasks; these include Graph WaveNet~\cite{wu2019graph}, AGCRN~\cite
 {bai2020adaptive}, and DMSTGCN~\cite{han2021dynamic}. However, these methods above are designed for either node-level or whole-graph tasks, and their abilities to extract
 important subgraphs have not yet been explored.

\textbf{Task-Specific Dynamic Graph Representation Learning.} Some studies have addressed the ``subgraph extraction'' problem and approached the solution through aggregation or denoising techniques. Titan~\cite{fu2019titan} considers the impact prediction on each road as an individual task and performs the prediction with a shared-parameter multitask model~\cite{fu2021hierarchical}. Meng~\etal~\cite{meng2022early} compare the performances of a CNN-based model and traditional traffic incident impact models. The results show that proper graph aggregation and kernel techniques improve the performance of dynamic graph learning models for traffic incident impact prediction tasks. Nevertheless, the three models above only consider the temporal impact of incidents and ignore the importance of spatial impact prediction.

As the survey above demonstrates, we can see that there has not been a formal
definition of the problem of spatiotemporal traffic incident impact prediction in the context of dynamic graph representation learning. Therefore, a new problem
definition, benchmark datasets, and directions for exploring subgraph impacts are required.

%% file: sections/problem_def.tex
\section{Problem Definition} In this section, we first define the traffic network graph and its expansion to the temporal dimension. Then, we quantify traffic incident impact with two measurements -- impact duration and impact length. Finally, we formulate the problem of incident impact prediction.

\subsection{Traffic Graph} A traffic performance measurement system usually sets static traffic loop sensors on the side of arterial roads and collects traffic data near the sensors. Previous research has tended to link groups of sensors that are closest to each other in the geographic or feature space. However, we argue that this assumption does not hold for freeways. For example, if two sensors are recording traffic in opposite directions on the same freeway, they could be as close to each other as the width of the road. However, these two sensors should not be linked, as a ``U'' turn is not an option on freeways. In addition, two sensors on different roads could also be geographically close at the intersection of two roads. In such cases, the real condition is that cars must drive through a long ramp to switch to another road, which means a much longer distance than the Euclidian distance. 

Considering that the relations are hard to quantify for two sensors on
different roads, we suggest constructing a dual-level graph to represent a
traffic network. The first level is the ``sensor-to-road'' level, which links sensors to the roads they are on. The second level (``road-to-road'' level) links two roads if they intersect. Note that two
directions on one freeway are considered two roads.
\begin{Definition}{\rm Dual-level traffic network graph.} Consider a traffic
 network graph as $G$, where $G=(S, R, E^{sr}, E^{rr}, E^{rs}, A^{sr}, A^
 {rr}, A^{rs})$. $S$ is the sensor node set of size $|S|$. $R$ is the road
 node set of size $|R|$. $E^{sr}, E^{rr}, E^{rs}$ are the edges linking
 sensor-road, road-road, and road-sensor, respectively. $A^{sr} \in \mathbb
 {R}^{|S|\times |R|}, A^{rr} \in \mathbb{R}^{|R|\times |R|}, A^
 {rs} \in \mathbb{R}^{|R|\times |S|}$ are the adjacent matrix form of $E^
 {sr}, E^{rr}, E^{rs}$.
\end{Definition}

Assume that $T$ timestamps around the incident validation time are used for prediction because they are usually most relevant to the incident impact. The traffic behaviors before and after an incident can be quite different. In this case, we split the $T$ timestamps into ``before validation'' (containing $T_{bv}$ timestamps) and ``after-validation'' (containing $T_{av}$ timestamps). 
\begin{Definition}{\rm Dynamic traffic network graph.} A dynamic traffic network graph can be denoted as $\mathcal{G}=\{\mathcal{G}_{bv}, \mathcal{G}_
 {av}\}=\{G_0, ..., G_{T_{bv}-1}, G_{T_{bv}}, ..., G_{T_{bv}+T_
 {av}-1}\}$. $G_0,...,G_{T_{bv}+T_
 {av}-1}$ indicates graphs at timestamp $0,...,T_{bv}+T_
 {av}-1$. $\mathcal{G}_{bv}$ and $\mathcal{G}_{av}$ are dynamic graphs before and after the validation time, respectively. 
\end{Definition}

\subsection{Incident Impact Prediction} The impact of an incident can be quantified into two dimensions: the spatial dimension and the temporal dimension. We represent the temporal impact with $\mathbf{Y}_{Dur}$, which is the difference between the validation time and the restoration time. This definition also matches the formal definition of incident duration provided by the Department of Transportation (DOT). For the spatial dimension, traditional transportation research counts the number of blocked cars upstream of the incident. However, cars’ speed can be affected by congestion even if they are not blocked. To account for this, we define the impact length $\mathbf{Y}_{Len}$ as the maximum continuous congestion road distance in the immediate upstream of the incident.
\begin{Definition}{\rm Incident impact precision.} Given a dynamic traffic
 network graph for an incident $\mathcal{G}$ and corresponding sensor feature
 tensor $\mathbf{X} \in \mathbb{R}^{|S|\times T \times C_{in}}$, the aim is
 to find a model $\mathcal{F}$ so that 
  \begin{equation}
 \mathcal{F}:(\mathcal{G}, \mathbf{X}) \rightarrow (\mathbf{Y}_{Dur}, \mathbf{Y}_{Len})
 \end{equation}
 where $C_{in}$ is the number of input channels recorded by the sensors, $Y_{Dur}$ is the impact duration, and $Y_{Len}$ is the impact length.
\end{Definition}

%% file: sections/method_copy.tex
\section{Methodology}
\label{sec:method} 
\begin{figure*}[h]
  \centering
  \includegraphics[width=\linewidth]{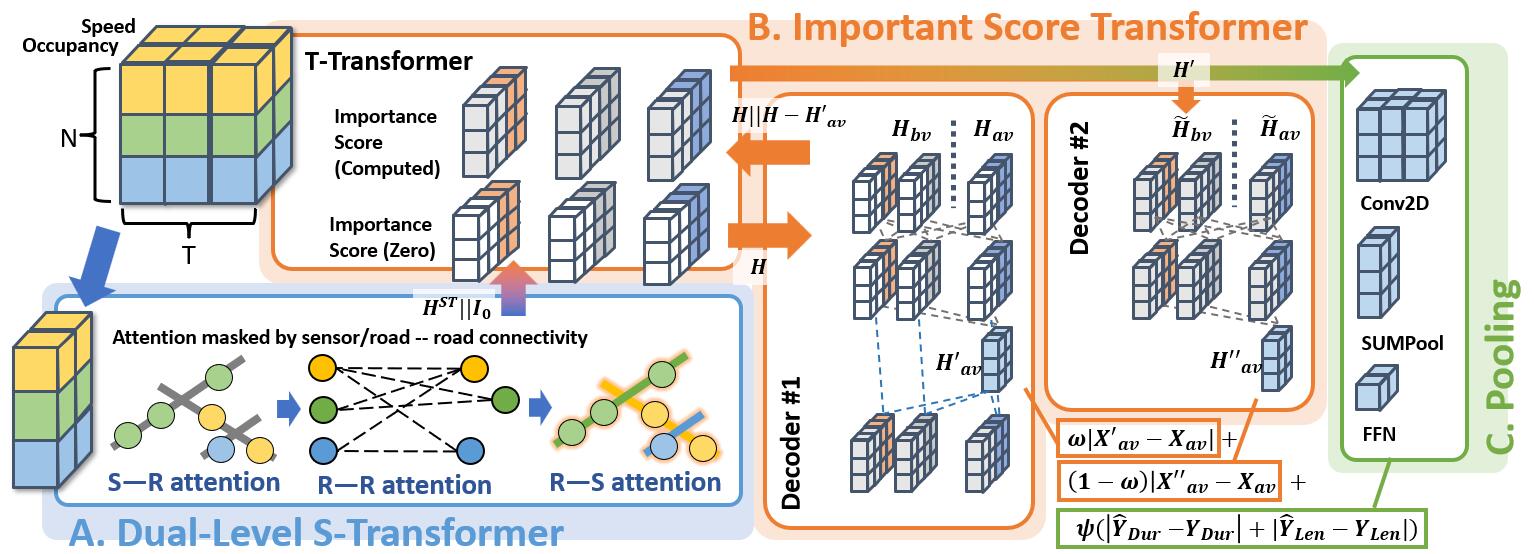}
  \caption{\textbf{The architecture of DG-Trans.} The blue rectangle labeled \textbf{A. Dual-Level Spatial Transformer} first encodes the input tensor by performing ``sensor-to-road,''  ``road-to-road,'' and ``road-to-sensor'' attentions. The orange rectangles labeled \textbf{B. Importance Score Transformer} further process the output of A $\mathbf{H}^{ST}$ with three modules.  $\mathbf{H}^{ST}$ is split into $\mathbf{H}_{bv}$ and $\mathbf{H}_{av}$ by the validation time of the incident, then fed to the T-Transformer with importance scores initialized as 0. The outputs $\mathbf{H}$ and $\mathbf{H}_{av}$ are sent to Decoder \#1 to reconstruct $\mathbf{X}_{av}$ and compute the importance score as $\mathbf{H}-\mathbf{H}'_{av}$. With the updated importance score, $\mathbf{H}^{ST}$ is fed to T-Transformer again. The output is further
   processed by Decoder \#2 to become $\mathbf{H}''_{av}$. The green rectangle labeled \textbf{C. Pooling} shows how the latent features are projected to the desired outputs. Finally, the loss is computed as the weighted combination of (1) the reconstruction from $\mathbf{H}'_{av}$ to $\mathbf{X}_{av}$, (2) the reconstruction from $\mathbf{H}''_{av}$ to $\mathbf{X}_{av}$, and (3) the prediction loss of impact duration and length.} % describe the architecture
  \label{fig:architecture}
\end{figure*}

As shown in Figure~\ref{fig:architecture}, the proposed architecture contains three main parts. The dual-level spatial transformer fuses spatial information among sensors for each timestamp. The importance score temporal transformer combines temporal information for each sensor, giving a higher ``importance score'' to sensors experiencing sudden feature changes after the validation time. A pooling module projects the learned dynamic graph representation to the expected output.

% modify the figure as the model changes

\subsection{Dual-Level Spatial Transformer} The design of the dual-level transformer is inspired by the concepts of anchored graphs and hypergraphs. Anchors have been used to reduce the complexity of attention in many studies~\cite
 {chen2020iterative, baek2021accurate, mialon2021graphit}. In our case, roads are considered anchor nodes for the sensors on them. While using anchors, the rank of the adjacency matrix decreases from $|S|$ to $|R|$. This means that multi-hop message-passing only matters among roads. The model maintains a global latent feature tensor $\mathbf{H}^{r} \in \mathbb{R}^{|R|\times T \times C}$ for each road at each timestamp, which does not vary by case. For each incident case, the sensor features are first used to update the case-irrelevant road feature in order to acquire case-relevant road features $\mathbf{H}^{sr}$. Then, the road features are further updated to $\mathbf{H}^{rr}$ by message-passing among roads. At this stage, the output road features can be considered both an intermediate pooling of the sensor features and a spatial feature fusion of the sensors. 
 % Therefore, $\mathbf{H}^{rr}$ is used for both impact prediction and sensor feature reconstruction. The second task is considered as a regularization of the first task so that the $\mathbf{H}^{rr}$ does not overfit the impact results or diverge from the raw representation of the sensor network.
 Finally, the road features are propagated back to the sensors for the next step.
 
 Denote sensors as $S$ and roads as $R$. A vanilla self-attention exploring spatial relations between sensors can be summarized as follows:
\begin{equation}
  a^{ss}_{ij} = \sigma(\frac{(\mathbf{Q} \mathbf{h}_{s_j})^T(\mathbf{K}\mathbf{h}_{s_i})}{\sqrt{d}})
\end{equation} 
where $s_i$ and $s_j$ are two different sensor nodes, $\mathbf{Q}$ and $\mathbf{K}$ are query and key projection parameters, $\mathbf{h}_{s_i}$ and $\mathbf{h}_{s_j}$ are embeddings of $s_i$ and $s_j$, $d=C \div n\_head$ is the dimension of each attention head, and $\sigma$ is the row-wise softmax activation function. 

In contrast to the vanilla approach, in S-Transformer, the attentive message-passing is accomplished with three transformer layers. The first transformer layer contains a ``sensor-to-road'' attention layer and a linear layer. The ``sensor-to-road'' attention computes the correlation
between sensors and roads and masks out the edge between $s_i$ and $r_j$ if
$s_i$ is not on $r_j$:
\begin{equation}
\label{eq:sr_attn}
\begin{aligned}
  &a^{sr}_{ij} = \sigma(mask_{sr}(\frac{(\mathbf{Q}^{sr}\mathbf{h}_{r_j})(\mathbf{K}^{sr}\mathbf{h}_{s_i})^T}{\sqrt{d}},m^{sr}_{ij})),\\
  &mask(\mathbf{x}, \lambda)=\begin{cases} \mathbf{x} & \lambda = 1 \\
                     -\infty &  \lambda = 0
       \end{cases}
\end{aligned}
\end{equation}
where $s_i$ and $r_j$ represent an arbitrary sensor and a road. $\mathbf{Q}^{sr}$ and
$\mathbf{K}^{sr}$ are query and key projection parameters for $r_j$ and $s_i$. $\mathbf{h}_{s_i}$ and $\mathbf{h}_{r_j}$ are embeddings of $s_i$ and $r_j$. $\mathbf{h}_{r_j} \in \mathbb{R}^{T\times C}$ is a
learnable parameter matrix. $\mathbf{M}^{sr} \in \{0, 1\}^{|S|\times|R|}$ is the
adjacent matrix between sensors and roads. $m^{sr}_{ij}=1$ if sensor $s_i$ is
on road $r_j$, otherwise $m^{sr}_{ij}=0$. $\mathbf{M}^{sr}$ performs as the mask of all the attention heads. With $a^{sr}_{ij}$, the road embedding can be updated as $\mathbf{h}_{r_j}^{sr}=\mathbf{W}^{sr}\sum_{i=1}^{|S|}a^{sr}_{ij}(\mathbf{V}^{sr}\mathbf{h}_{r_j})+\mathbf{b}^{sr}$.

The attention in the second transformer is a ``road-to-road'' self-attention layer intended to
extract road intersection information. The attention can be expressed as
follows:
\begin{equation}
  a^{rr}_{ij} = \sigma(mask_{rr}(\frac{(\mathbf{Q}^{rr}\mathbf{h}_{r_j}^{sr})(\mathbf{K}^{rr}\mathbf{h}_{r_i}^{sr})^T}{\sqrt{d}},\mathbf{M}^{rr}_{ij}))
\end{equation}
where $\mathbf{M}^{rr} \in \{0, 1\}^{|R|\times|R|}$ represents four levels of adjacency between
roads: $m^{rr}_{ij}=1$ if (1) $i=j$, (2) $r_i$ intersects $r_j$, (3) $r_i$ intersects $r_k$ and $r_j$ intersects $r_k$, and (4) fully connected. The four different masks can be applied to different attention heads. In our design, each of the four masks was applied to one attention head. A spatial-relation awared road feature tensor is then computed as $\mathbf{h}_{r_j}^{rr}=\mathbf{W}^{rr}\sum_{i=1}^{|R|}a^{rr}_{ij}(\mathbf{V}^{rr}\mathbf{h}_{r_j}^{sr})+\mathbf{b}^{rr}$.

The last transformer contains a ``road-to-sensor'' attention, meaning that sensor vectors are queries, and road vectors are keys and values. The output of this step propagates the aggregated road features to the sensors. 
% is used to examine if $\mathbf{H}^{rr}$ can represent the raw sensor network. 
The equation is essentially the reversed version of Equation~\ref
{eq:sr_attn}:
\begin{equation}
  \mathbf{H}^{rs} = \sigma(\frac{(\mathbf{Q}^{rs}\mathbf{H}^{rr})(\mathbf{K}^{rs}\mathbf{H}^{rr})^T}{\sqrt{d}})(\mathbf{V}^{rs}\mathbf{H}^{rr})
\end{equation}

Note that there is no mask in this step, which preserves the attention
mechanism's flexibility. Previous works have usually linked the top-k closest sensors in
the geographic or feature space, weighting the links with the distances.
However, we observe that attention is good at learning weights but weak at
learning graph structures. In this case, we simply need to find nodes that
are definitely linked to each other and leave the weight learning task to
attention. Therefore, we chose to partially control the graph structure with unweighted adjacent matrices. Finally, we applied skip-connection, layer-normalization, and dropout to $\mathbf{H}^{rs}$.

Essentially, S-Transformer stresses the effects of sensors on
high-degree-centrality roads. The strengths of the S-Transformer can be
summarized as follows: (1) it avoids the human error introduced by manually
choosing ``road-to-road'' for ``top-k'', (2) it allows long-range message-passing as all sensors on intersected roads are linked, (3) it preserves the flexibility of attention with a relatively small number of edges
($|S|+|R|^2+|S||R|, |R|\ll |S|$ at most) (4) it is more time efficient ($\mathcal{O}(|S||R|^2+|R|^3+|R|^2|S|), |R|\ll |S|$) than traditional attention mechanisms ($\mathcal{O}(|S|^3)$) and requires fewer layers as the spatial message-passing is performed sufficiently by the ``road-to-road'' self-attention module.

\subsection{Importance Score Transformer} As an incident usually affects only
 a relatively small part of the whole traffic network, treating all sensors equally for prediction may introduce unwanted noise. However, manually extracting sensors near the incident may cause the loss of long-range complex impact
 patterns and even the most representative features due to the early/delayed response of the traffic network. In this case, a method that dynamically locates the region and time window affected by the incident is important. Based on the assumption that the traffic measurements of sensors affected by incidents show more obvious changes than other sensors, we locate the incidents with anomaly detection techniques (i.e., assigning sensors with larger variance a higher ``importance score''). 

Inspired by~\cite{tuli2022tranad}, our importance score
transformer contains three modules: a temporal transformer
(T-Transformer) and two decoders. The T-Transformer module encodes the output of the S-Transformer with and without the importance score along the time dimension. The first decoder computes the importance score and reconstructs
$\mathbf{X}_{av}$, while the second reconstructs $\mathbf{X}_{av}$ from the combination of the importance score and the graph embedding. Denote the ``after-validation'' section of $\mathbf{H}^{ST}$ as $\mathbf{H}_{av}$ and the ``before-validation'' section of $\mathbf{H}^{ST}$ as $\mathbf{H}_{bv}$. The combination of the T-Transformer and any one of the decoders is equivalent to a classic transformer network when considering $\mathbf{H}^{ST}$ as the input sequence and $\mathbf{H}_{av}$ as the output sequence.

Assume the importance score is $\mathbf{I} \in \mathbb{R}^
{|S|\times T \times C}$ and the output of S-Transformer as $\mathbf{H}^{ST} \in \mathbb{R}^{|S|\times T \times C}$. The output of the T-Transformer can be written as follows:
\begin{equation}\label{eq:score_1}
  \mathbf{H} = \text{TTrans}([\mathbf{H}^{ST} || \mathbf{I}_0])
\end{equation}
where $\text{TTrans}()$ indicates T-Transformer, which is a block sequentially performing temporal self-attention and skip-connection. $\mathbf{I}_0$ is the initialized importance score (which is an all-zero tensor).

The task for both Decoder \#1 and Decoder \#2 is to reconstruct $\mathbf{X}_{av}$. Each decoder has a self-attention layer and a mutual-attention layer. 
Decoder \#1 attempts to achieve its goal using $\mathbf{H}$ and $\mathbf{H}_{av}$:
\begin{equation}
  \mathbf{H}'_{av} = \text{mu-attn}_1(\mathbf{H}, \mathbf{H}, \text{self-attn}_1(\mathbf{H}_{av}))
\end{equation}

The three parameters in $\text{mu-attn}_1$ are placeholders for value, key, and query.

Then, the importance score is updated as $\mathbf{I} = \mathbf{H}^{ST} - \mathbf{H}'_{av}$. Note that $\mathbf{H}^{ST}$ has $T$ timestamps while $\mathbf{H}'_{av}$ has
$T_{av}$ timestamps. We examined various methods, such as repeating timestamps in $\mathbf{H}'_{av}$ and getting mean/min/max of $\mathbf{H}'_{av}$ along the time axis. All the methods resulted in similar performance. Accordingly, we
adopt the general form to represent the difference between $\mathbf{H}^{ST}$ and $\mathbf{H}'_{av}$. 
% We directly use the difference between $\mathbf{H}^{ST}$ and $\mathbf{H}'_
% {av}$ as we want to distinguish the traffic patterns before and after the
% incident occurrence/ Intuitively, for affected sensors, the average speed before the incident is larger than $avg(\mathbf{H}_{bv}) \geq \mathbf{H}'_{av}$ and
% $\mathbf{H}_{av} \leq \mathbf{H}'_{av}$ based on the intuition of anomaly
% detection. 
Replacing $\mathbf{I}_0$ with $\mathbf{I}$, we apply the T-Transformer and Decoder \#2 the same way as the previous steps, with the concatenated $\mathbf{H}^{ST}$ and $\mathbf{I}$ as the input. 

\begin{equation}\label{eq:score_2}
\begin{aligned}
  &\mathbf{H}' = \text{TTrans}([\mathbf{H}^{ST} ||\mathbf{I}])\\
  &\mathbf{H}''_{av} = \text{mu-attn}_2(\mathbf{H}', \mathbf{H}', \text{self-attn}_2(\mathbf{H}_{av}))
\end{aligned}
\end{equation}

Finally, $\mathbf{H}'_{av}$ and $\mathbf{H}''_{av}$ are used to reconstruct $\mathbf{X}_{av}$ separately with the same two-layer feed-forward network (FFN):
\begin{equation} \label{eq:ffn}
\begin{aligned}
  &\mathbf{X}'_{av} = \mathbf{W}_{1,2}(\phi( \mathbf{W}_{1,1}\mathbf{H}'_{av}+ \mathbf{b}_{1,1}))+ \mathbf{b}_{1,2} \\
  &\mathbf{X}''_{av} = \mathbf{W}_{1,2}(\phi( \mathbf{W}_{1,1}\mathbf{H}''_{av}+ \mathbf{b}_{1,1}))+ \mathbf{b}_{1,2}
\end{aligned}
\end{equation}
where $\mathbf{W}_{1,1}, \mathbf{b}_{1,1}, \mathbf{W}_{1,2}, \mathbf{b}_{1,2}$ are parameters of the two linear layers and $\phi$ is the activation function, which is the LeakyReLu in the FFN in Equation~\ref{eq:ffn}. 

\subsection{Pooling and Loss} After the spatial and temporal encoding, the spatiotemporal representation $\mathbf{H}'$ is considered well-learned and ready for prediction. To further refine the features, the importance score is used to weight the elements in $\mathbf{H}'$ through element-wise multiplication. This is followed by a temporal dimension aggregation through a 2D convolution layer and a spatial dimension aggregation through SUMPooling. Lastly, three linear layers with LeakyReLU activations are used to project the features into the desired dimension. This process can be summarized as follows:
\begin{equation}
\begin{aligned}
  &\mathbf{H}'_T = \text{Conv2d}(\mathbf{H}'), \mathbf{H}'_T \in \mathbb{R}^{|S|\times C}\\
  &\mathbf{H}'_S = \text{SUMPool}(\mathbf{H}'_T), \mathbf{H}'_T \in \mathbb{R}^C)\\
  &\hat{Y} = \mathbf{W}_{p3}\phi(\mathbf{W}_{p2}\phi(\mathbf{W}_{p1}\mathbf{H}'_S+\mathbf{b}_{p1})+\mathbf{b}_{p2}))+\mathbf{b}_{p3}
\end{aligned}
\end{equation}
where $\hat{Y}\in \mathbb{R}^{C_{out}}, C_{out}=2$. The first value represents the impact duration, and the second is the impact length. SUMPooling is chosen because it is the most expressive pooling method~\cite{xu2018powerful}, and we want the pooled representation to be capable of representing all possible incident scenarios. L1 loss is chosen as the prediction loss function so that the model focuses on the overall trends of $\mathbf{Y}$ instead of outliers. The prediction loss can then be written as $Loss_1 = |\mathbf{\hat{Y}}_{Dur}-\mathbf{Y}_{Dur}|+|\mathbf{\hat{Y}}_{Len}-\mathbf{Y}_{Len}|$.

The second part of the loss is the reconstruction loss. This loss is for regularization purposes and is self-supervised. The only objective of the model is to predict $\mathbf{Y}_{Dur}$ and $\mathbf{Y}_{Len}$. L1 loss is employed for both $\mathbf{X}'_{av}$ and $\mathbf{X}''_{av}$, i.e., the loss functions are $Loss_2 = |\mathbf{X}'_{av}-\mathbf{X}_{av}|$ and $Loss_3 = |\mathbf{X}''_{av}-\mathbf{X}_{av}|$.
To generate the importance scores correctly, the Importance Score Transformer is trained in an adversary way. Here, we explain our design in terms of a GAN in order to make it understandable. Consider $\mathbf{H}_{av}$ as the true "image", $\mathbf{H}_{bv}$ as the fake "image", Decoder \#1 as the discriminator, and Decoder \#2 as the generator. In the first several epochs, the weight of $Loss_2$ is far larger than the weight of $Loss_3$. As a result, Decoder \#1 discriminates $\mathbf{H}_{bv}$ and $\mathbf{H}_{av}$ better by assigning $\mathbf{H}_{av}$ larger attentive weights in $\text{TTrans}()$. As the weight of $Loss_3$ increases, Decoder \#2 is trained to "generate" a new $\mathbf{H}_{bv}$ by concatenating it with $\mathbf{I}_{bv} \approx \mathbf{H}_{bv}-\mathbf{H}_{av}$. The new $\mathbf{H}_{bv}$ -- $\mathbf{H}'_{bv}$ recieves more attention in $\text{TTrans}()$ with the importance score and thus enlarge $Loss_2$. This way, as the weight of $Loss_3$ grows, the attentive weights in $\text{TTrans}()$ finally stabilize at some point that slightly inclines to $\mathbf{H}_{bv}$, which makes Decoder \#1 produce a larger importance score for $\mathbf{H}_{av}$ and a smaller score for $\mathbf{H}_{bv}$.

The loss during training can be written as follows:
\begin{equation} \label{eq:loss}
  Loss = \psi Loss_1 + \omega Loss_2 + (1-\omega) Loss_3
\end{equation} 
where $\psi$ is the weight of the prediction loss. $\omega \in (0, 1]$ is the weight of the reconstruction loss of
Decoder \#1, while $(1-\omega)$ is the weight of the reconstruction loss of Decoder \#2. $\omega$ is initialized as $1$ and decreases as the number of epochs increases. 

%% file: sections/experiment_copy.tex
\section{Experiment}
\label{sec:experiment}
In this section, we first introduce our datasets and our method for data
cleansing and labeling. Then, we compare DG-Trans with state-of-the-art
baselines and evaluate our proposed model from three perspectives: prediction
precision, module-level effectiveness, and execution efficiency of the
spatial attention module. Next, we perform a case study to examine whether DG-Trans can correctly extract spatiotemporal features. Finally, we propose several basic methods of incorporating new data into DG-Trans. These methods may help to broaden the problem domain of traffic incident impact prediction.

\subsection{Dataset} 
The traffic loop sensor data used for this research are collected from the Caltrans Performance Measurement System (PeMS)\footnote{\url{https://pems.dot.ca.gov/}}; The incident record data are from RITIS \footnote{\url{https://www.ritis.org/}}; The road networks are
downloaded from Tiger Priscroads \footnote{\url
{https://www2.census.gov/geo/tiger/}}. Based on the location of sensors and incidents, we selected several freeway segments in Los Angeles and San Bernardino regions, which have well-constructed and complex road networks. 

In addition to the data used in this research (i.e., sensor measurements, adjacency information, and incident impact records), we also provided auxiliary data to broaden the potential problem domain. Specifically, our datasets
contain three basic elements: roads, sensors, and incidents. For roads, we provide a matrix indicating whether two roads intersect each other and the location of the intersections (in the form of the distance from the start point of the road). For sensors, we collect their positions on the roads and
their five-minute speed and occupancy measurements. As the rate of missing records is less than 0.005\%, we simply filled those values with daily average
speed/occupancy. Finally, for incidents, we provide the position on the road,
the DateTime (number of five minutes from 2019/09/01 00:00:00), the incident
category, the impact duration, and the impact length. 

Since we lacked the ground truth of impact length, we acquired the label with three
steps. We first extracted the regular weekly traffic pattern by averaging the
speeds from 2019/08/01 to 2019/10/31. In this case, a speed lower than the
regular pattern is considered an indicator of non-recurrent congestion. The second step is to split the impacted sensor measurements from the others. We performed a binary 1D k-means classification to extract low speed measurements, then looked for the closest upstream sensor that detected no
congestion during the incident clearance process. If the selected sensor is the closest sensor to the incident, the impact length is zero. Otherwise, the impacted length is defined by the distance from the incident to the upstream sensor before the picked sensor.

\begin{figure}
\centering
\begin{subfigure}{.49\linewidth}
    \centering
    \includegraphics[width=\linewidth]{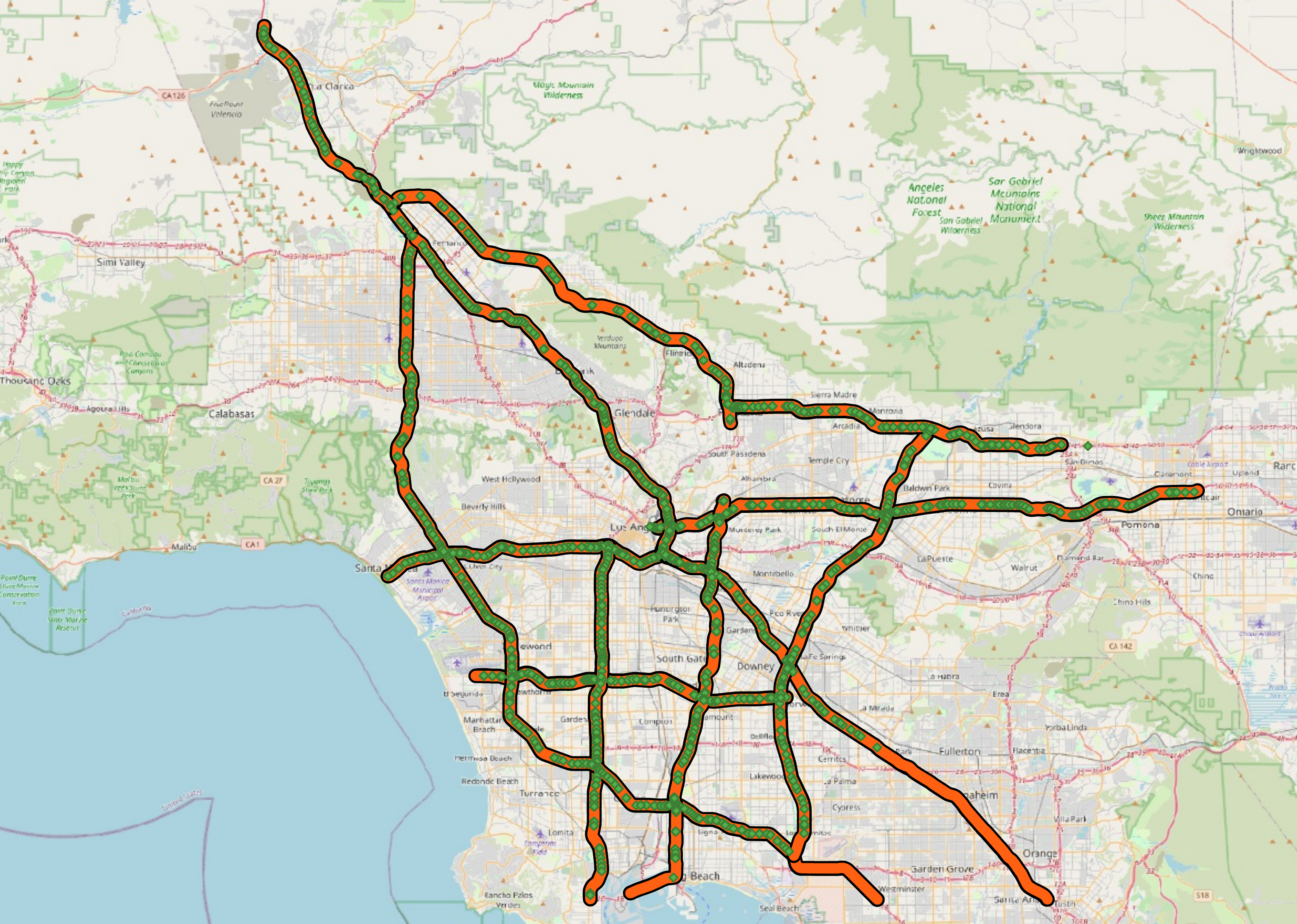}
    \caption{Included freeways with incident locations (Incident-LA)}\label{fig:p7_road}
\end{subfigure}
\begin{subfigure}{.49\linewidth}
    \centering
    \includegraphics[width=\linewidth]{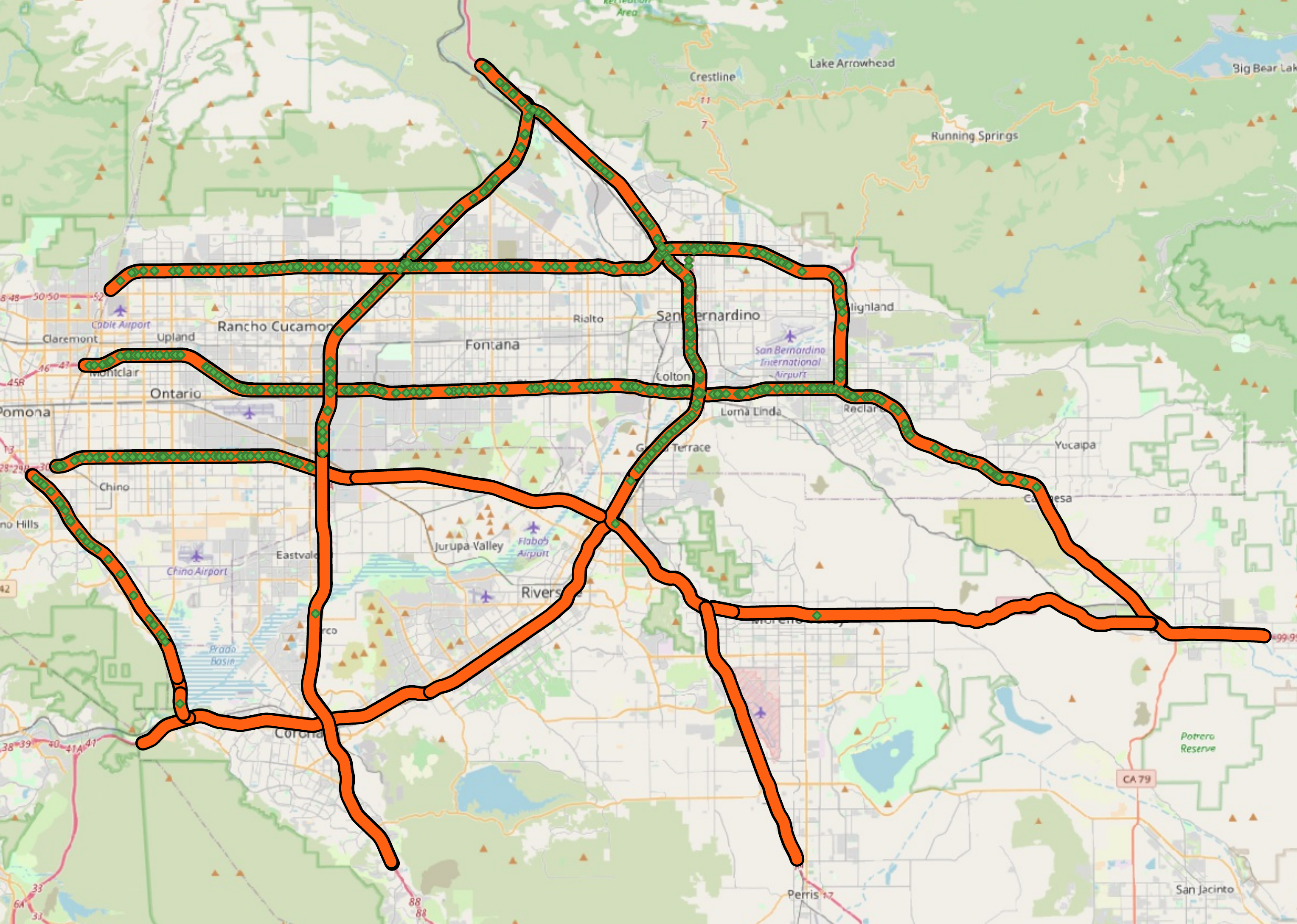}
    \caption{Included freeways with incident locations (Incident-SB)}\label{fig:p8_road}
\end{subfigure}
\bigskip
\begin{subfigure}{\linewidth}
    \centering
    \includegraphics[width=\linewidth,height=.4\linewidth]{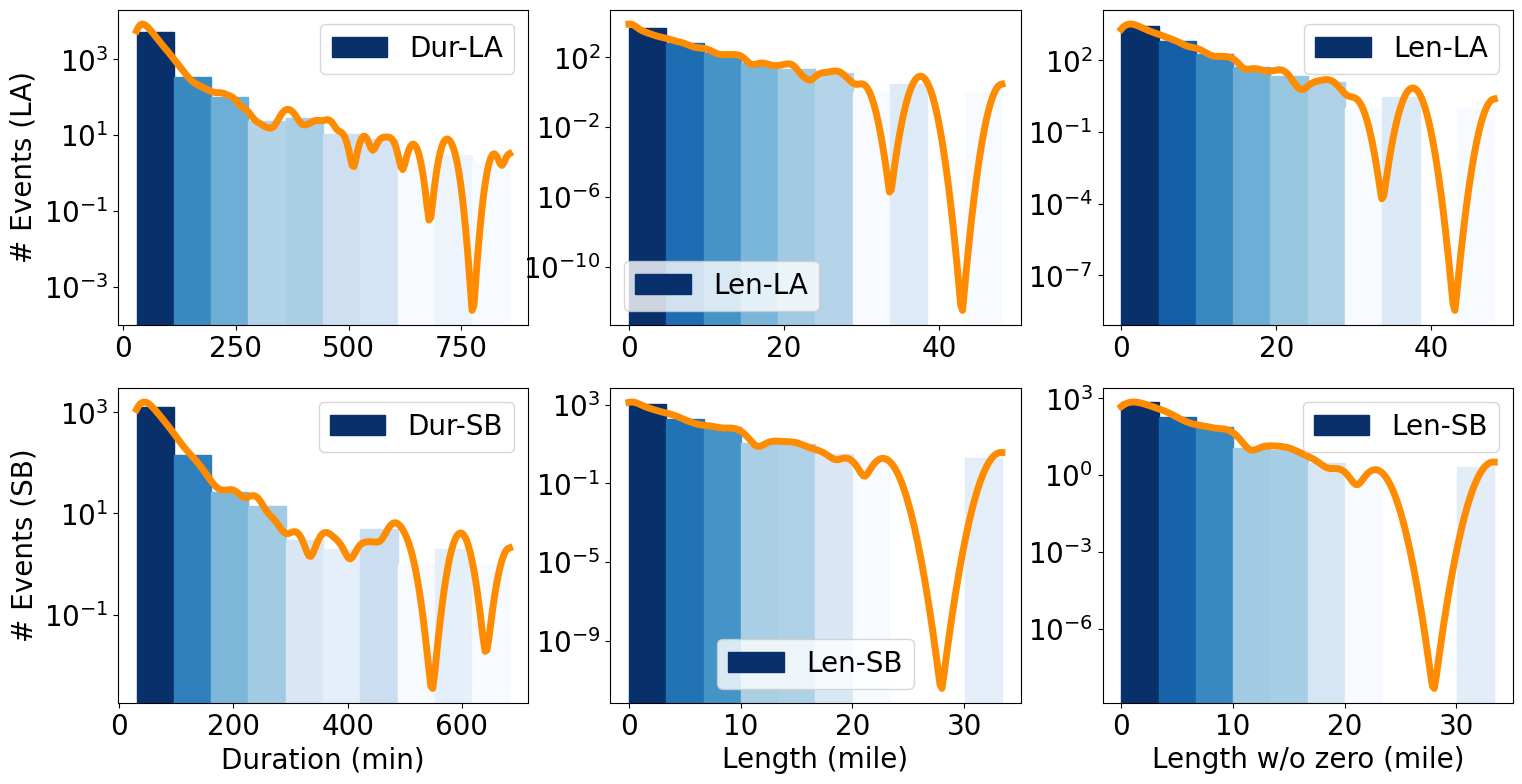}
    \caption{Distribution of event duration/impact length/ non-zero impact length (Incident-SB)}\label{fig:p8_dis}
\end{subfigure}
\caption{Basic Dataset Information}
\label{fig:dataset_intro}
\end{figure}

To construct properly scaled datasets, we selected several inter-state freeways as illustrated in Figure~\ref{fig:p7_road} and~\ref{fig:p8_road}. The time span was set to one month (2019/09/01 -- 2019/09/30). Sensors and incidents not on the chosen freeways were filtered out. We also removed incidents with a duration of fewer than 30 minutes due to their limited temporal impact. Figure~\ref{fig:p8_dis} shows the distribution of impact duration and length. The orange dashed lines indicate the fluctuations of log event counts across different durations and impact lengths. The shades of the bars show the magnitude of event counts. All the label values follow power-law distributions except that
the impact length in the San Bernardino region is relatively noisy compared
with the others. Finally, we present two traffic incident impact
prediction datasets: Incident-LA and Incident-SB. The detailed attributes of
the two dynamic networks are illustrated in Table~\ref{tab:pems_dist}. 

% \begin{figure*}
% \centering
% \begin{subfigure}{.33\linewidth}
%     \centering
%     \includegraphics[width=\linewidth]{figures/pems07.pdf}
%     \caption{Covered Freeways with Incident Locations (Incident-LA)}\label{fig:p7_road}
% \end{subfigure}
%     \hfill
% \begin{subfigure}{.65\linewidth}
%     \centering
%     \includegraphics[width=.99\linewidth,height=.4\linewidth]{figures/distribution_PEMS-07_1.png}
%     \caption{Distribution of Event Duration/Impact Length/ Non-Zero Impact Length (Incident-LA)}\label{fig:p7_dis}
% \end{subfigure}
% \bigskip
% \begin{subfigure}{.33\linewidth}
%     \centering
%     \includegraphics[width=\linewidth]{figures/pems08.pdf}
%     \caption{Covered Freeways with Incident Locations (Incident-SB)}\label{fig:p8_road}
% \end{subfigure}
%    \hfill
% \begin{subfigure}{.65\linewidth}
%     \centering
%     \includegraphics[width=.99\linewidth,height=.4\linewidth]{figures/distribution_PEMS-08_1.png}
%     \caption{Distribution of Event Duration/Impact Length/ Non-Zero Impact Length (Incident-SB)}\label{fig:p8_dis}
% \end{subfigure}
% \caption{Basic Dataset Information}
% \label{fig:dataset_intro}
% \end{figure*}

\begin{table}[h]
\small
  \centering
  \begin{tabular}{cllll}
    \toprule
    Dataset & \# event & \# node & \# edge & \# 0 length \\
    & & (R/S) & (R-R/S-S) & \\
    \midrule
    Incident-LA & 5,668 & 32/1,663 & 142/869,640 & 2,062\\
    Incident-SB & 1,452 & 28/1,150 & 140/390,822 & 454\\
    
  \bottomrule
\end{tabular}
  \caption{Dataset Properties}
  \label{tab:pems_dist}
\end{table}

\subsection{Baselines} To evaluate the efficiency of DG-Trans, nine representative models are chosen to perform the same task. Considering that only a few models target similar tasks, we choose the two most recently
 published incident impact prediction models, two conventional models, and
five state-of-the-art traffic forecasting models. Among the latter five models, two leverage attentive graph representation learning techniques, while the other three use graph convolution.
\begin{itemize}
    \item \textbf{L-1 regularized linear regression (LASSO)~\cite{tibshirani1996regression}.} As LASSO only takes one-dimensional features as inputs, we examined different feature aggregation methods. These include (1) averaging over both the spatiotemporal dimensions, (2) picking the closest upstream/downstream sensor and averaging over the temporal dimension, and (3) picking the closest upstream/downstream sensor and the five minutes immediately after the validation time. For parameter selection, we examined $\lambda$ values of $0.1$, $1.0$, $10.0$, and $100.0$. 
    \item \textbf{Support vector regression (SVR)~\cite{tibshirani1996regression}.} Similar to LASSO, we examined three different feature aggregation methods. We used the default parameters ($C=1, \epsilon=0.1$) in the sklearn~\cite{scikit-learn} package of Python.
    \item \textbf{HastGCN~\cite{fu2021hierarchical}.} HastGCN is a spatiotemporal attention model designed for incident duration prediction. We reproduce the model with the help of the author. Some features used in~\cite{fu2021hierarchical} are removed because they are not provided in our datasets. All the other settings are the same as the original model.
    \item \textbf{AGWN~\cite{meng2022early}.} AGWN preprocesses the adjacency matrix with a wavelet filter before the graph convolution operation. The model utilizes only one timestamp immediately after the validation time. To allow a fair comparison, we attached our Importance Score Transformer and pooling module to it. All the hyperparameters are the same as in the original paper.
    \item \textbf{STTN~\cite{xu2020spatial}.} STTN utilizes transformers for both spatial and temporal message-passing. The model is designed for node-level speed  forecasting. In this case, we attach our pooling module to the original STTN. The experiment code is extracted from the STTN github repository \footnote{\url{https://github.com/wubin5/STTN}}. All the hyperparameters remain the same as is mentioned in the paper.
    \item \textbf{STAWnet~\cite{tian2021spatial}.} STAWnet utilizes attentive graph message-passing and gated temporal convolution networks (TCN). It is also a node-level speed forecasting model and requires our pooling module to fit our incident impact prediction task. We utilized the official code \footnote{\url{https://github.com/CYBruce/STAWnet}} and left the hyperparameters unchanged.
    \item \textbf{DMSTGCN~\cite{han2021dynamic}.} DMSTGCN decomposes the adjacency matrix into four trainable embeddings for graph convolution. The temporal information is merged with gated TCN. Our pooling module was attached to the model to obtain the desired output. All the hyperparameters are the same as in the original paper \footnote{\url{https://github.com/liangzhehan/DMSTGCN}}.
    \item \textbf{Graph WaveNet~\cite{wu2019graph}.} Graph WaveNet contains a self-adaptive graph convolution module and a dilated temporal convolution module. Our pooling module was attached to the model to obtain the desired output. All the hyperparameters are the same as in the original paper \footnote{\url{https://github.com/nnzhan/Graph-WaveNet}}.
    \item \textbf{AGCRN~\cite{bai2020adaptive}.} AGCRN performs node-adaptive graph convolution and GRU-like temporal message-passing. Our pooling module was attached to the model to obtain the desired output. All the hyperparameters are the same as in the original paper \footnote{\url{https://github.com/LeiBAI/AGCRN}}.
\end{itemize} 

\subsection{Hyperparameter and Metrics}
In the problem settings of this paper, we assume that the objective is to predict the incident's impact within a short time after the event. To do so, nine timestamps (six for ``before-validation'' and three for ``after-validation'') were adopted. As a result, the model could not see the full traffic pattern during the incident.

For the training process of DG-Trans, we adopted a batch size of 8 due to the GPU memory limitation. The learning rate was 0.0005 with a 0.001 weight decay. The number of attention heads was 4. The LeakyReLU factor was set to 0.2, and the dropout rate was 0.1. In the loss function (Equation~\ref{eq:loss}), the prediction loss weight $\psi$ was 1.0, and $\omega$ was equal to the epoch.

Following our previous works~\cite{fu2019titan, fu2021hierarchical,meng2022early}, we adopted root mean squared error (RMSE) and mean absolute
error (MAE) as metrics. However, the impact length introduces labels of zero
value making mean absolute percentage error (MAPE) invalid. Therefore, we replaced MAPE with symmetric mean absolute percentage error (sMAPE). Based
on the definition ($RMSE=\sqrt{\frac{1}{N}\sum_{i=1}^N(y_i-\hat
{y}_i)^2}$, $MAE=\frac{1}{N}sum_{i=1}^N|y_i-\hat{y}_i|$, $sMAPE=\frac{1}
{N}sum_{i=1}^N|\frac{2|y_i-\hat{y}_i|}{|y_i|+|\hat{y}_i|}|$), RMSE penalizes
large gaps more harshly than MAE, while sMAPE focuses more on the magnitude of the differences from the true values.

\begin{table*}[t]
  \small
  \centering
  \begin{tabular}{c lll lll lll lll lll}
    \toprule
    &\multicolumn{3}{c}{Incident-LA (dur (min))} & 
     \multicolumn{3}{c}{Incident-LA (len (mile))} &
     \multicolumn{3}{c}{Incident-SB (dur (min))} &
     \multicolumn{3}{c}{Incident-SB (len (mile))}\\
     & RMSE & MAE & sMAPE & RMSE & MAE & sMAPE & RMSE & MAE & sMAPE & RMSE & MAE & sMAPE\\
    \midrule
    
    LASSO~\cite{tibshirani1996regression}&  59.773& 51.776& 0.760& 8.270& 6.794& 1.211& 58.921& 51.263& 0.757& 10.743& 8.934& 1.112\\
    SVR~\cite{tibshirani1996regression}& 60.073& 50.763& 0.743& 8.559& 6.300& 1.299& 59.560& 50.924& 0.761& 11.632 & 8.812 & 1.218\\
    HastGCN~\cite{fu2021hierarchical} & 31.719& 20.372& 0.319& 8.421& 6.402& 1.272& 35.936& 25.078& 0.381& 13.053& 9.299& 1.350 \\
    AGWN~\cite{meng2022early} & 31.934& 20.720& 0.341& 10.840& 6.594& 0.874& 32.864& 22.391& 0.365& 11.730& 8.736& 0.743 \\
    
    STTN~\cite{xu2020spatial}& 31.826& 21.098& 0.322& 9.644& 6.317& 0.893& 31.235& 20.631& 0.326& 12.346& 9.025& 0.812 \\
    STAWnet~\cite{tian2021spatial}& 31.400& 20.315& 0.318 & 8.619
& 6.311& 1.310 & 29.280& 20.212& 0.320& 11.994& 8.945& 1.260\\  
    DMSTGCN~\cite{han2021dynamic}& 31.555 & 20.342 & 0.319 & 10.638 & 7.784 & 0.880 & 29.810 & 20.263 & 0.312 & 12.929 & 9.846 & 0.791 \\
    Graph WaveNet~\cite{wu2019graph} & 31.880&20.415&0.320&10.489&7.672&0.861& 30.765&20.455&0.324& 14.093&10.807& 0.914 \\
    AGCRN~\cite{bai2020adaptive} & 31.253&20.363&0.319& 8.662&6.212&0.899& 30.905&20.652&0.324&12.808&9.093&1.172 \\
    \midrule
    \textbf{DG-Trans} & \textbf{31.413}& \textbf{20.310}& \textbf{0.318}& \textbf{9.494}& \textbf{6.226}& \textbf{1.477}& \textbf{29.726}& \textbf{20.140}& \textbf{0.319}& \textbf{11.731}& \textbf{8.818}& \textbf{1.235} \\
    \bottomrule
  \end{tabular}
  \caption{\textbf{RMSE, MAE, and sMAPE for Duration and Impact Length Prediction of Incident-LA and Incident-SB.} This table lists the performance of nine state-of-the-art baselines and our proposed model.}
  \label{tab:baseline}
\end{table*}

\subsection{Prediction Result Analysis}
The input to the models includes the adjacent matrices and sensor measurements one hour before and half an hour after the validation time. The output is the impact duration and length. We evaluated the duration and length separately as they are of different units.

Table~\ref{tab:baseline} illustrates the performance of DG-Trans against the baselines. 

\textbf{Conventional baselines.} For LASSO and SVR, the best performance is
 achieved with the closest upstream sensor and the first timestamp after the
 validation time as inputs. The results appeared to be insensitive to
 hyperparameters. As shown in Table~\ref{tab:baseline}, even though they
 produced competitive results in RMSE and MAE impact length prediction, LASSO
 and SVR performed poorly in predicting impact duration prediction.  This result matches our hypothesis that the ``closest'' sensors and timestamps are not optimal for prediction. 

\textbf{Spatiotemporal neural network baselines.} DG-Trans surpasses the other baselines (i.e., spatiotemporal neural networks) in most of the metrics and achieves competitive performance on the other metrics if it is not the best. Specifically, DG-Trans outperforms another spatiotemporal transformer, STTN, by about 10\% in duration prediction and 5\% in length prediction. DG-Trans also beats previous incident impact prediction models, HastGCN and AGWN, by approximately 10\% in performance.

 We find that DG-Trans cannot beat the other models in all metrics. To prove the efficiency of our model, we rank all ten models by each criterion and list the average rank in Table~\ref{tab:rank}. The table shows that our model performs the best on average.

 %  This part of the experiment shows that models designed specifically for
 % incidents (i.e., focus on locating incidents and denoising) perform better
 % than models evenly treat every spatiotemporal feature.
 \begin{table}[h]
\small
  \centering
  \begin{tabular}{clllll}
    \toprule
    Model&\textbf{DG-Trans}&STTN&STAW.&DMST.&AGCRN\\
    \midrule
    Rank & \textbf{3.58}&3.75&4.08&5.25&5.25\\
    \midrule
    Model &AGWN&G.W.&LASSO&Hast.&SVR\\
    \midrule
    Rank &5.75&6.67&6.83& 6.92&6.92\\
    
  \bottomrule
\end{tabular}
  \caption{\textbf{Average rank of baseline and DG-Trans performance.} Abbreviated model names are used due to the limited space. The smaller the number, the higher the average rank.}
  \label{tab:rank}
\end{table}

\subsection{Ablation Study}

\begin{table*}[t]
  \small
  \centering
  \begin{tabular}{c lll lll lll lll lll}
    \toprule
    &\multicolumn{3}{c}{Incident-LA (dur (min))} & 
     \multicolumn{3}{c}{Incident-LA (len (mile))} &
     \multicolumn{3}{c}{Incident-SB (dur (min))} &
     \multicolumn{3}{c}{Incident-SB (len (mile))}\\
     & RMSE & MAE & sMAPE & RMSE & MAE & sMAPE & RMSE & MAE & sMAPE & RMSE & MAE & sMAPE\\
    
    \midrule
    No-STrans & 31.245& 20.320& 0.319& 9.608& 6.251& 1.503& 28.793& 21.240& 0.338& 11.922& 8.877& 1.255 \\
    No-TTrans & 31.674& 20.371& 0.319& 9.496 & 6.215& 1.570& 32.024& 27.265& 0.422& 12.432& 8.903& 1.344 \\
    No-Road & 31.611& 20.351& 0.319& 9.909& 6.345& 1.616& 28.568& 20.928& 0.333& 13.240& 9.286& 1.496 \\
    No-Score & 31.846& 20.321& 0.319& 9.496& 6.215& 1.473& 30.000& 20.236& 0.320& 14.000& 9.743& 1.688 \\

    \midrule
    \textbf{DG-Trans} & \textbf{31.413}& \textbf{20.310}& \textbf{0.318}& \textbf{9.494}& \textbf{6.226}& \textbf{1.477}& \textbf{29.726}& \textbf{20.140}& \textbf{0.319}& \textbf{11.731}& \textbf{8.818}& \textbf{1.235} \\
    \bottomrule
  \end{tabular}
  \caption{\textbf{RMSE, MAE, and sMAPE for Duration and Impact Length Prediction of Incident-LA and Incident-SB.} The results of the ablation study include our model without the S-Transformer (No-STrans), our model without the T-Transformer (No-TTrans), our model without the road anchors (No-Road), and our model without the importance score (No-Score).}
  \label{tab:ablation}
\end{table*}

We conducted four ablation experiments to evaluate the contributions of each component of the DG-Trans model. The result is shown in Table~\ref{tab:ablation}. For the ``No-STrans'' and ``No-TTrans'' experiments, we simply removed the corresponding modules. For ``No-Score'', we skipped the concatenation steps in Equation~\ref{eq:score_1} and ~\ref{eq:score_2} and removed the reconstruction losses during training. For ``No-Road'', we replaced the Dual-Level S-Transformer with the transformer used in STTN. Initially, we assumed that ``No-Road'' would outperform DG-Trans as our graph had too many edges removed. However, the results in Table~\ref{tab:ablation} show that our model cannot achieve its performance without any of its components. We observed that the performance drops less in ``No-Score'' and ``No-TTrans'' than in ``No-Road'' and ``No-STrans''. This may be because the number of timestamps in the input is too small for a transformer to work.  
% \subsection{Graph Attention Efficiency.} 

% We also compared the execution time
%  of S-Transformer with the same modules in the baselines. Based on our
%  observation, HastGCN leverages three linear transformations to perform the
%  attention while STTN and STAWnet utilize the vanilla attention mechanism. As
%  suggested by Table~\ref{tab:pems_efficient}, S-Transformer executes $2$ to
%  $5$ times faster than the vanilla attention and also faster than the linear
%  transformation attention. 

% The execution times of STAWnet on the two datasets are the same, as we cut off
% some nodes from Incident-LA to remove the OOM issue. We also observe that the
% execution time is not linearly related to the number of sensors in DG-Trans.
% The reason is that the number of computations is also related to the number
% of roads. As the numbers of roads are similar in the two datasets, the
% execution times are the same as well.

% \begin{table}[h]
% \small
%   \centering
%   \begin{tabular}{cllll}
%     \toprule
%     Dataset & HastGCN & STTN & STAWnet & \textbf{DG-Trans} \\
%     \midrule
%     Incident-LA & 0.0021 & 0.0050 & 0.0080 & \textbf{0.0015}\\
%     Incident-SB & 0.0019 & 0.0025 & 0.0080  & \textbf{0.0013}\\
    
%   \bottomrule
% \end{tabular}
%   \caption{Average execution time of the graph attention module in second}
%   \label{tab:pems_efficient}
% \end{table}

\subsection{Case Study} 
\begin{figure}[h]
  \centering
  \includegraphics[width=\linewidth]{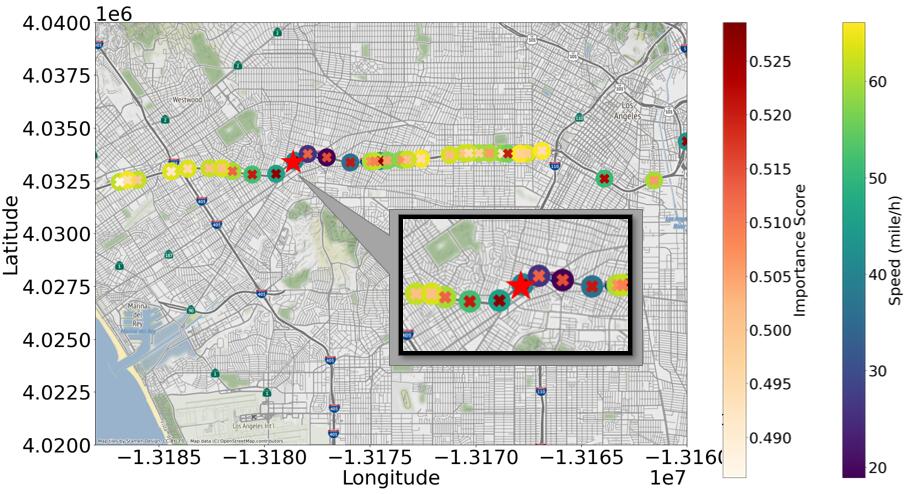}
  \caption{\textbf{Case study of incident \#18693 on the Freeway I--10.} This is the map of an incident and the surrounding traffic loop sensors on I--10 in the Incident-LA dataset. The red star indicates the location of the incident, and the yellow--green--blue dots are sensors colored according to speed measurements. The orange--red ``X''s indicate the importance scores assigned by DG-Trans. } % describe the architecture
  \label{fig:score}
\end{figure}

To examine whether our importance score truly helps to
 identify incidents, we explored several incident cases. One example is
 incident \#18693 (Figure~\ref{fig:score}). The map plots the
 five-minute average speed immediately after the validation time and the importance
 scores within the same time slot. It is obvious that sensors that detect
 lower speeds also have higher importance scores, while high importance
 scores cluster around the incident (red star). However, we also observe that
 incident-irrelevant speed drops lead to high importance scores as well. This is why the model utilizes both the score and the embedded features for prediction.

\noindent
\textbf{Beyond This Task.} We believe that the other parts of the dataset,
 (i.e., the data not used in this paper), can be used to increase the
 prediction accuracy. We examined simple methods of merging the
 auxiliary information into the traffic network, such as adding an incident
 classification task, using road and sensor position for position encoding,
 and embedding incident metadata. While none of those methods worked, all of these attempts are also uploaded to our GitHub repository \footnote{\url{https://github.com/styxsys0927/DG-Trans.git}} for others to investigate.

%% file: sections/conclusion.tex
\section{Conclusion}
\label{sec:conclusion}
In this paper, we introduce two traffic incident impact datasets and give a
definition of incident impact prediction tasks in the context of
spatiotemporal data mining. We design a new
transformer-based impact prediction model, which contains a novel dual-level spatial transformer module and an importance score temporal transformer module to assist in identifying
measurements affected by the incidents. In addition, we evaluated the performance of two conventional models and six deep graph traffic forecasting models on our datasets. The experiments show that DG-Trans outperforms the other models. Moreover, the ablation studies show that the dual-level attention strategy outperforms general attention layers, and the case study shows that the importance score transformer is able to identify incident-relevant sensors. 